\documentclass[10pt,twocolumn,letterpaper]{article}

\usepackage{cvpr}              

\usepackage{graphicx}
\usepackage{amsmath}
\usepackage{amssymb}
\usepackage{booktabs}
\usepackage[table]{xcolor}
\definecolor{Graylight}{gray}{0.9}
\usepackage{colortbl}
\usepackage[utf8]{inputenc} 
\usepackage[T1]{fontenc}    
\usepackage{hyperref}       
\usepackage{url}            
\usepackage{booktabs}       
\usepackage{amsfonts}       
\usepackage{nicefrac}       
\usepackage{microtype}      
\usepackage{multirow}
\usepackage{comment}
\usepackage{color}

\usepackage[ruled,vlined]{algorithm2e}
\usepackage{caption}

\newlength\savewidth
\newcommand{\tablestyle}[2]{\setlength{\tabcolsep}{#1}\renewcommand{\arraystretch}{#2}\centering\footnotesize}
\renewcommand{\paragraph}[1]{\vspace{1.25mm}\noindent\textbf{#1}}

\newcolumntype{x}[1]{>{\centering\arraybackslash}p{#1pt}}
\newcolumntype{y}[1]{>{\raggedright\arraybackslash}p{#1pt}}
\newcolumntype{z}[1]{>{\raggedleft\arraybackslash}p{#1pt}}
\definecolor{baselinecolor}{gray}{.9}

\definecolor{commentcolor}{RGB}{110,154,155}   
\usepackage[capitalize]{cleveref}
\crefname{section}{Sec.}{Secs.}
\Crefname{section}{Section}{Sections}
\Crefname{table}{Table}{Tables}
\crefname{table}{Tab.}{Tabs.}

\def\cvprPaperID{322} 
\def\confName{CVPR}
\def\confYear{2023}

\begin{document}

\title{MaskCLIP: Masked Self-Distillation Advances Contrastive \\ Language-Image Pretraining}
\author{
Xiaoyi Dong$^{1}$\thanks{Equal contribution, $\dagger$ Corresponding Author}  \thanks{Work done during an internship at Microsoft Research Asia}, Jianmin Bao$^{2*}$, Yinglin Zheng$^{3}$, Ting Zhang$^{2}$,  Dongdong Chen$^{4,\dagger}$, Hao Yang$^{2}$, \\Ming Zeng$^{3}$, Weiming Zhang$^{1}$, Lu Yuan$^{4}$, Dong Chen$^{2}$, Fang Wen$^{2}$, Nenghai Yu$^{1}$ \\
$^{1}$University of Science and Technology of China 
$^{2}$Microsoft Research Asia \\
$^{3}$Xiamen University
$^{4}$Microsoft Cloud + AI \\
{\tt\small\{dlight@mail., zhangwm@, ynh@\}.ustc.edu.cn} 
{\tt\small cddlyf@gmail.com}\\
{\tt\small\{jianbao, ting.zhang, luyuan, doch, fangwen\}@microsoft.com } \\
{\tt\small \{zhengyinglin@stu., zengming@\}xmu.edu.cn} 
{\tt\small yanghao.alexis@foxmail.com}
}
\maketitle

\begin{abstract}
This paper presents a simple yet effective framework MaskCLIP, which incorporates a newly proposed masked self-distillation into contrastive language-image pretraining. The core idea of masked self-distillation is to distill representation from a full image to the representation predicted from a masked image. Such incorporation enjoys two vital benefits. First, masked self-distillation targets local patch representation learning, which is complementary to vision-language contrastive focusing on text-related representation.
Second, masked self-distillation is also consistent with vision-language contrastive from the perspective of training objective as both utilize the visual encoder for feature aligning, and thus is able to learn local semantics getting indirect supervision from the language.
We provide specially designed experiments with a comprehensive analysis to validate the two benefits. 
Symmetrically, we also introduce the local semantic supervision into the text branch, which further improves the pretraining performance.
With extensive experiments, we show that MaskCLIP, when applied to various challenging downstream tasks, achieves superior results in linear probing, finetuning, and zero-shot performance with the guidance of the language encoder. Code will be release at \url{https://github.com/LightDXY/MaskCLIP}.
\end{abstract}

\section{Introduction}
Vision-language (VL) contrastive learning~\cite{radford2021learning,jia2021scaling} has shown remarkable success in pretraining for various tasks. With large-scale image-text pairs available on the Internet, the model composed of a simple dual encoder design learns strong semantic prior by aligning between image and text. The resulting visual encoder not only exhibits excellent linear probing and finetuning performance, but also enables impressive zero-shot performance with the guidance of the language encoder, showing the generality of natural language and its ability to supervise a wide range of visual concepts.

Nonetheless, the associated language description, though providing richer information than mere class labels, still can hardly describe all the information in the corresponding image, as images are continuous signals with fine-grained details and complex semantics. As a result, the VL contrastive by aligning global representations may only focus on the text-described objects and ignore the rest which might be useful for downstream tasks.

In this paper, we are interested in how to fully leverage the image itself to facilitate the VL contrastive to further improve the transfer capability.
(1) Firstly, the learned feature representation shall characterize local patches, serving as a complementary for global representation in VL contrastive.
Inspired by the recent success of masked image modeling~\cite{he2021masked,bao2021beit,wang2022bevt,radford2021learning,dong2021peco,wei2021masked} in learning patch representations, we also randomly mask the input image with a large portion to force the visual encoder to focus on the remaining visible patches.
(2) Secondly, the learned representation for local patches shall possess semantic meanings,
being consistent with the global representation receiving semantic text supervision.
We bring mean teacher self-distillation~\cite{tarvainen2017mean,grill2020bootstrap} to supervise the learned patch representations with the visual feature representations, enabling implicit supervision from natural language.
The resulting objective is denoted as \emph{masked self-distillation} where the student model and the teacher model come from the same neural networks and the knowledge is distilled from the full image (fed to the teacher model) to the masked image (fed to student model).
To this end, we introduce MaskCLIP by incorporating masked self-distillation into VL contrastive to advance the transferable visual encoder.

\begin{figure*}
\centering
\includegraphics[width=1\linewidth]{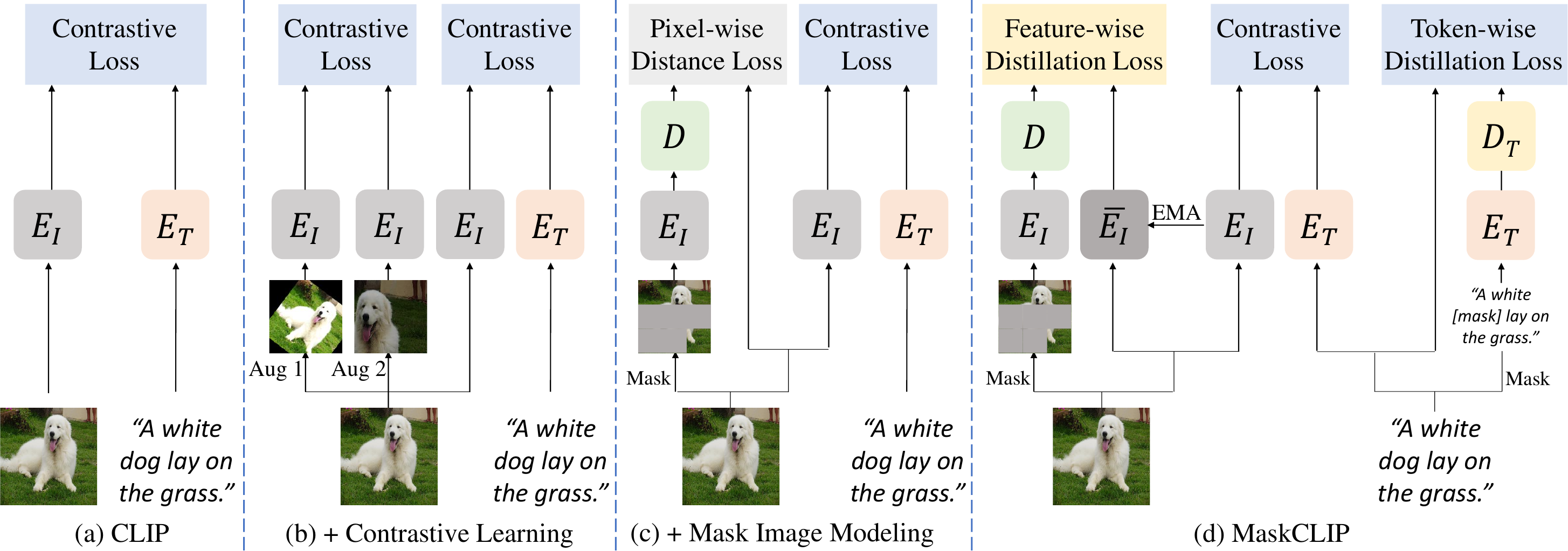}
\vspace{-5mm}
\captionof{figure}{Pipeline comparison between combination CLIP with different vision self-supervised learning methods. (a) Vanilla CLIP. (b) CLIP + contrastive learning. (c) CLIP + pixel prediction mask image modeling. (d) CLIP + mask self-distillation, \ie MaskCLIP. 
The $E_T$, $E_I$ is the text encoder and image encoder respectively, and all the $E_I$, $E_T$ within each pipeline share the weight.
$\bar{E}_I$ is the mean-teacher model, whose weight is updated by the exponential moving average of $E_I$ and does not require gradient. }
\label{fig:framework}
\vspace{-4mm}
\end{figure*}

There are several recent attempts~\cite{mu2021slip,zheng2021general} also exploring the capability of the visual encoder under natural language supervision.
The common approach is to introduce contrastive learning or masked image modeling on the vision side together with contrastive language-image pretraining. However, the performance indeed improves based on CLIP but does not as well as our masked self-distillation.
We argue that (1) the contrastive learning objective based on central crop augmentation actually learns global representations for salient objects while lack of attention on the surrounding backgrounds~\cite{chen2022context}; and (2) masked image modeling usually needs to remap the learned representation to pixels~\cite{he2021masked} or discrete tokens~\cite{bao2021beit}. Such low-level prediction target is inefficient for semantic feature learning and thus also conflicts with high-level language supervision in VL contrastive. A brief illustration is presented in Figure~\ref{fig:framework}. In the experiments, we conduct comprehensive ablations to analyze the difference and provide numerical and visual evidence for better understanding.

Symmetrically, we argue that local semantic supervision on the text branch is also helpful for the text encoder and eventually beneficial for zero-shot performance. So we introduce the same mask-data-modeling format supervision into the text branch as well. 
Different from images where the pixel is low-level signal, the words crafted by human beings are already highly semantic, so we use the tokenized word piece as the prediction target directly, following the well-studied mask language modeling method BERT. Meanwhile, to reduce the output conflicts between contrastive learning and mask language modeling, we introduce a small decoder for the mask language modeling branch.

We train our MaskCLIP on a subset of a publicly available image-text pairs dataset, YFCC \cite{thomee2016yfcc100m}, and thoroughly evaluate the transfer ability of visual representations on several vision benchmarks: ImageNet-1K~\cite{deng2009imagenet} for classification, ADE20K~\cite{zhou2017scene} for semantic segmentation, MS-COCO~\cite{lin2014microsoftcoco} for detection and segmentation, as well as a batch of other classification benchmarks. 
When it comes to ImageNet-1K~\cite{deng2009imagenet} classification, MaskCLIP achieves $+6.9\%$, $+7.2\%$, $+1.3\%$ higher than CLIP for zero-shot transfer, linear probing, and finetuning respectively. For vision downstream tasks, we reach $+2.7$ mIoU on ADE20K~\cite{zhou2017scene} and $+1.8$ AP$^b$, $+1.4$ AP$^m$ on MS-COCO~\cite{lin2014microsoftcoco}. For vision-language tasks, MaskCLIP achieves $+6.1\%$ average zero-shot accuracy on 20 datasets, and $+17.2\%$, $+12.8\%$ rank@1 improvement on the Flickr30K~\cite{young2014image} image-test retrieval.
In the recent Image Classification in the Wild challenge academic track, our MaskCLIP gets the $1_{st}$ result with $48.9\%$ TOP-1 average accuracy, surpassing the second team with $3.4\%$.

In summary, the major contributions of this work are:
\begin{enumerate}
	\item We present a novel vision-language pretraining framework MaskCLIP, by introducing masked self-distillation objective to facilitate VL contrastive for better transferable visual models.
	\item We present extensive ablation studies on MaskCLIP variants and provide in-depth analysis numerically and visually to help understand how the proposed masked self-distillation assists VL contrastive.
	\item We demonstrate our MaskCLIP on tens of benchmarks, showing the superiority under all three settings: zero-shot, linear probing, and finetuning.
\end{enumerate}

\section{Related Work}

\noindent\textbf{Vision-language pretraining} 
Recent years have seen rapid progress made in vision-language pretraining~\cite{li2019visualbert,lu2019vilbert,tan2019lxmert,su2019vl,chen2020uniter,li2020unicoder,li2020oscar,zhou2020unified,lu202012,desai2021virtex,li2020hero,luo2020univl,qi2020imagebert,li2020unimo,li2021align}. Several multiple cross-modality loss functions have been proposed for the training objective, such as 
image-text matching~\cite{li2019visualbert,lu2019vilbert,tan2019lxmert,chen2020uniter,xu2021e2e}, masked language modeling~\cite{li2019visualbert,lu2019vilbert,tan2019lxmert,su2019vl,chen2020uniter}, masked image modeling~\cite{lu2019vilbert,tan2019lxmert,su2019vl,chen2020uniter}, contrastive loss~\cite{li2020oscar,li2020unimo,li2021align}. These objects are often mixed with each other to form a compound objective. While a variety of approaches have been proposed, few works investigate the performance on visual representation learning for image classification. Recently, CLIP~\cite{radford2021learning} and ALIGN~\cite{jia2021scaling} show that the image-text contrastive learning objective achieves promising performance for visual representation learning. 
There are many following works proposed to further improve the pretraining performance, DeCLIP~\cite{zhou2021denseclip}, SLIP~\cite{mu2021slip}, COTS~\cite{lu2022cots}, ViCHA~\cite{shukor2022efficient}, CYCLIP~\cite{goel2022cyclip} use additional uni/multi-modality supervision to improve the model capability, and PyramidCLIP~\cite{gao2022pyramidclip}, KLITE~\cite{shen2022k}, IDEA~\cite{huang2022idea} seek to external knowledge from pre-trained models or datasets as the additional guidance. FILIP~\cite{yao2021filip} and LOUPE~\cite{li2022fine} introduce fine-grained alignment to the model. 
Focusing on this research direction, we analyze the desired properties of supervision which could be complementary to CLIP, and propose the masked self-distillation objective incorporated with the image-text contrastive loss to further improve pretraining performance for various visual understanding tasks.

\noindent\textbf{Self-supervised learning}
Self-supervised visual representation learning has attracted increasing attention over the past few years. The objective of the self-supervised learning is mainly divided into two categories: contrastive and generative~\cite{liu2021self}. The contrastive methods, such as MOCO~\cite{he2019moco,chen2020mocov2}, SimCLR~\cite{chen2020simple,chen2020big}, BYOL~\cite{grill2020bootstrap}, SimSiam~\cite{chen2020exploring}, and DINO~\cite{caron2021emerging} measure the similar and dissimilar samples by contrastive loss. Their success heavily depends on the strong data augmentation. The generative methods, such as BEiT \cite{bao2021beit}, MAE \cite{he2021masked}, PeCo \cite{dong2021peco}, BEVT \cite{wang2022bevt}, BootMAE~\cite{dong2022bootstrapped} and MaskFeat \cite{wei2021masked} leverage masked image modeling to reconstruct the remaining masked part of its original input from the given visible parts. The generative methods show more promising transfer performance than the contrastive methods,
as generative objective learns patch representations while contrastive objective focuses on learning centric global representations~\cite{chen2022context}.

\noindent \textbf{Self-knowledge distillation}
Self-knowledge distillation~\cite{kim2020self} aims to distill the knowledge in a model itself and uses it for training the model. Instead of distilling knowledge from a pretrained teacher model~\cite{hinton2015distilling}, self-knowledge distillation regards a temporal ensemble of the student model as the teacher. It means that a student model becomes a teacher model itself, which gradually utilizes its own knowledge for softening the hard targets to be more informative during training. Self-knowledge distillation has been explored in semi-supervised learning~\cite{tarvainen2017mean}, contrastive learning~\cite{cheng2021data,li2021align}, self-supervised learning~\cite{baevski2022data2vec,caron2021dino}. 
In this paper, we use visual features supervised by natural language for guidance in masked self-distillation which naturally fit VL contrastive to learn more transferable visual representations.

\section{MaskCLIP}

We introduce MaskCLIP, a novel framework that learns visual representations.
The core part of MaskCLIP is its backbone image encoder, denoted by $E_I$ as shown in Figure \ref{fig:framework}. It obtains the transferable capability during pretraining that could benefit downstream vision tasks.
Following recent self-supervised approaches \cite{he2021masked,mu2021slip,bao2021beit,chen2021empirical},
we implement the backbone $E_I$ as a Vision Transformer (ViT) \cite{dosovitskiy2020image}. 
The prediction results from $E_I$ given an input image $I$ then should be a collection of visual feature tokens, represented as 
\begin{equation}
E_I(I) = \{f_\textit{cls}, f_1, f_2, \dots, f_N\}.
\label{equ:encode}
\end{equation} 
Here \emph{cls} is short for class token. $1, \dots, N$ are the indexes of the non-class tokens.

The rest of this section starts with the utilization of language supervision. More shall be emphasized on the masked self-distillation, which we deem crucial for visual pretraining.

\subsection{Vision-language Contrastive}

Following \cite{jia2021scaling,radford2021learning}, we introduce a Transformer-based text encoder $E_T$ to leverage language knowledge. 
It aims to align the global feature representations of an image and a text with respect to some forms of similarity.
Precisely, consider a given image-text pair $\{I, T\}$, besides extracting the visual feature representation $E_I(I)$ using the vision backbone as shown by Equation \ref{equ:encode}, we additionally use the text encoder $E_T$ to extract linguistic features from the text $T$. 

The mean feature of the two branches are regarded as the global representations and are fed into a projection head (implemented as a fully-connected layer) respectively to obtain the metric embeddings $e^T$ and $e^I$.
Image-text contrastive loss is employed to align them during pretraining. The loss can be formulated as $\mathcal{L}_T + \mathcal{L}_I$, with
\begin{align}
\mathcal{L}_I = -\frac{1}{B}\sum_{i=1}^B\log\frac{\exp(e^I_i e^T_i / \sigma)}{\sum_{j=1}^B \exp(e^I_i e^T_j / \sigma)} \nonumber\\
\mathcal{L}_T = -\frac{1}{B}\sum_{i=1}^B\log\frac{\exp(e^T_i e^I_i / \sigma)}{\sum_{j=1}^B \exp(e^T_i e^I_j / \sigma)},
\end{align}
where $B$ stands for the number of image-text pairs within a training mini-batch, $i, j$ are indexes within the batch; $\sigma$ stands for the temperature for the loss functions, which is learned together with all other parameters during training.

\subsection{Masked Self-distillation for Visual Encoder}

Knowledge distillation is a learning paradigm where a student model is trained to match the output of a given teacher model, so that the student model can be improved by the teacher.
Instead of bringing in an external teacher, self-distillation methods such as \cite{tarvainen2017mean,caron2021dino,grill2020bootstrap} proposes using a \emph{mean teacher} model that is derived from the student itself. 
In specific, the teacher shares the same structure with the student, while the parameters of the teacher are exponential moving averages (EMA) of the parameters from the student. In the following, we would use the term ``EMA model'' to represent such mean teacher model constructed from the student.

MaskCLIP leverages the mean teacher self-distillation to enhance its vision representations.
Let $\bar{E}_I$ be the EMA model of the backbone encoder $E_I$.
$\theta_t$ and $\bar{\theta}_t$ are the parameters of $E_I$ and $\bar{E}_I$ at training step $t$. $\bar{\theta}_t$ is updated with
\begin{equation}
\bar{\theta}_t = \alpha \bar{\theta}_{t-1} + (1-\alpha) \theta_t,
\label{equ:ema}
\end{equation}
where $\alpha$ is a hyper-parameter for smoothing updates.
We propose to incorporate masked image modeling into self-distillation, resulting in \emph{masked self-distillation} with asymmetric input for student model and teacher model.

In specific, considering a given input image $I$, we first feed it to the EMA model $\bar{E}_I$ (teacher model) to obtain the distillation targets. These target features can be represented as 
\begin{equation}
\bar{E}_I(I)=\{\bar{f}_\textit{cls}, \bar{f}_1, \bar{f}_2, \dots, \bar{f}_N\}.
\label{equ:distill_target}
\end{equation}
In the meantime, we randomly mask a large portion of the input image patches and then feed it into the original backbone $E_I$ (student model).
Following \cite{he2021masked}, we only feed the visible (unmasked) patches, denoted by $I'$, into the original backbone $E_I$ to speed up computation and save memory. Let $\mathcal{M}$ be the indexes of all the masked tokens. These encoded features corresponding to visible tokens can then be denoted as $E_I(I')=\{f'_\textit{cls}\}\bigcup\left\{f'_{k \not\in \mathcal{M}}\right\}$.
They are then joined with a shared and learnable feature vector, denoted as $m$, that represents mask tokens, to form a complete set of features $\{f'_\textit{cls}, f'_1, f'_2, \dots, f'_N\}$, with $f'_{i \in \mathcal{M}}=m$. 
We attach positional embeddings onto all these tokens, and append a small Transformer $D$ as a decoder to predict features of the masked region from the visible tokens,
which could be formulated as
\begin{align}
(D \circ E_I)(I') & = D\left(\left\{  f'_\textit{cls}, f'_1, f'_2, \dots, f'_N   \right\}\right) \nonumber\\
 & = \left\{  f''_\textit{cls}, f''_1, f''_2, \dots, f''_N   \right\}.
\end{align}
Inspired by ~\cite{zhou2021ibot}, we use an online quantizer $h()$ to transform the output features into a soft codewords distribution, and minimize the cross-entropy between the target features and the predicted features. Formally, 
\begin{equation}
\mathcal{L}_\textit{Dist} = \frac{1}{| \mathcal{M} |} \sum_{k \in \mathcal{M}} - \bar{h}(\bar{f}_k)^T log\, h(f''_{k}).
\label{equ:distill_loss}
\end{equation}
here the parameter of the teacher quantizer $\bar{h}()$ is also EMA updated by the online quantizer, similar to the teacher model.

\subsection{Local Semantic Learning for Text Encoder}
Besides the local semantic supervision for the visual encoder, we argue it is also helpful for the text encoder. So we introduce the BERT pretraining into the text branch. For the text $T=\{t_{sos},t_1, t_2,...,t_M,t_{eos}\}$, we denote the masked input as $T'=\{t_{sos}',t_1', t_2',...,t_M',t_{eos}'\}$, where $t_{i \in \mathcal{M}_T}'=m_t$ and $t_{i \notin \mathcal{M}_T}'=t_i$, and $\mathcal{M}_T$ be the indexes of all the masked text tokens. 
The output feature of the encoder is $E_T(T')$. 

To reduce the output conflict between the global image-text contrastive learning and the local mask language modeling, we further introduce a small text decoder, which shares the same architecture as the encoder but with only a few layers. So that the global prediction and local prediction are conducted at different layers. We denote the output feature as:
$(D_T \circ E_T)(T') = \{t_{sos}'',t_1'', t_2'',...,t_M'',t_{eos}''\}$
and the loss could be formulated as:
\begin{equation}
\mathcal{L}_\textit{MLM} = \frac{1}{| \mathcal{M}_T |} \sum_{k \in \mathcal{M}_T} - t_k^T log\, t''_{k}.
\label{equ:distill_loss}
\end{equation}

\subsection{Overall Loss Functions}

Finally, we pretrain MaskCLIP with all these losses combined: 
\begin{align}
	\mathcal{L}_I+\mathcal{L}_T+\lambda \mathcal{L}_\textit{Dist} +\beta \mathcal{L}_\textit{MLM},
\end{align} 
with $\lambda, \beta$ being the hyper-parameter weighting between VL contrastive loss and self-supervised learning loss.
All the components of MaskCLIP are trained from scratch, including the visual backbone $E_I$, the visual decoder $D$, the text encoder $E_T$, as well as the text decoder $D_T$.

\begin{figure*}
\centering
\includegraphics[width=0.98\linewidth]{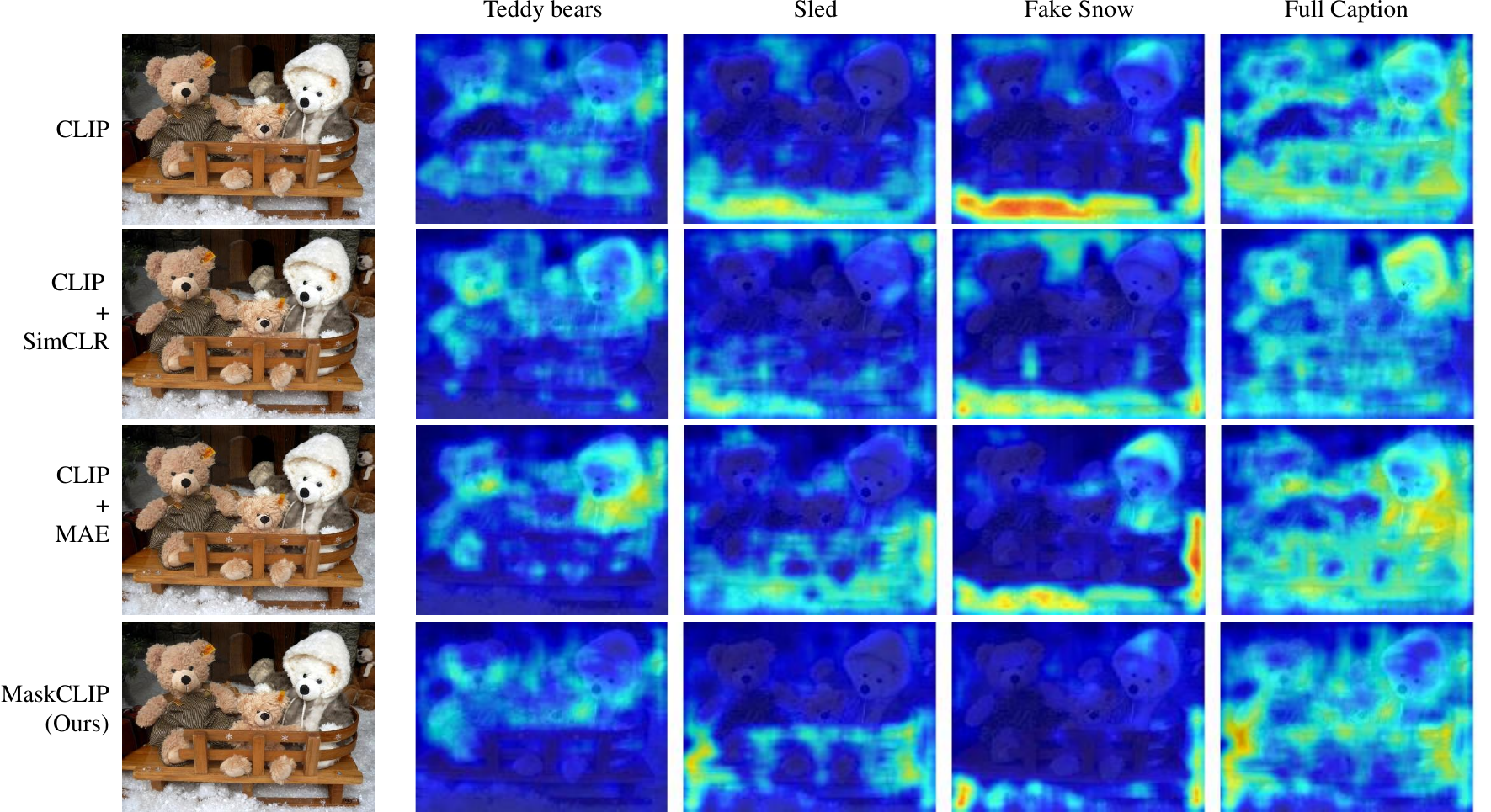}
\vspace{-3mm}
\captionof{figure}{Visualization of the similarity between text and image features. The images and captions are from the MS-COCO val set. Here we show the image feature similarity with both full caption and different objects in it. The caption is ``Three teddy bears sit in a sled in snow''. More results could be found in the supplemental materials.}

\label{fig:vis}
\vspace{-3mm}
\end{figure*}

\section{Experiments}
\subsection{Setup}
\noindent\textbf{Model architecture.}
Our framework consists of the visual encoder $E_I$, the text encoder $E_T$, the visual decoder $D$, and the text decoder $D_T$. We adopt the widely used Transformer ViT-B/16~\cite{dosovitskiy2020image} for a fair comparison. It is composed of 12 layers, 768 width, and 12 head. The input image is $224\times 224$ resolution and is further split into $14\times 14$ patches with size $16\times 16$. A learnable cls token is prepended to the 196 embeddings.
For the text encoder, we adopt a 12-layer, 512-width, and 8-head Transformer following CLIP~\cite{radford2021learning}, and the text decoder has 4 layers.
The number of text tokens is fixed to 77 with necessary truncations or paddings.
For the image decoder, we directly use a one-layer Vision Transformer.

\noindent\textbf{Pretraining details.}
We train our proposed MaskCLIP from scratch for 25 epochs, the batch size is fixed to 4096 for all the experiments. The masks used in the mask self-distillation branch and mask language modeling branch are random mask with a mask ratio of 75\% and 20\%. 
We pretrain all the models with the commonly used YFCC15M dataset, which is flited from the YFCC100M~\cite{thomee2016yfcc100m} dataset by ~\cite{radford2021learning}. 

\noindent\textbf{Downstream details.}
We evaluated MaskCLIP on several downstream datasets, including ImageNet-1K~\cite{deng2009imagenet}, ADE20K~\cite{zhou2017scene}, MS-COCO~\cite{lin2014microsoftcoco}, Flicr30K~\cite{young2014image} \etal
For ImageNet-1K, we report zero-shot, linear probing, and finetuning performance. The zero-shot is conducted following the label prompt setting in SLIP~\cite{mu2021slip}. For linear probing, we fix the backbone and train a new linear classifier for 90 epochs. 
For finetuning, we follow the setting in BEiT ~\cite{bao2021beit} and finetune the model for 100 epochs with a layer-decayed learning rate. 
See supplemental materials for more details.

\begin{table}[t]
\small

\centering
\setlength{\tabcolsep}{0.4 mm}{
\begin{tabular}{l|cc|ccc|cc}
\toprule
\multirow{2}{*}{} & \multicolumn{2}{c|}{Training}  & \multicolumn{3}{c|}{IN-1K} & \multicolumn{2}{c}{Flicker30K}\\
 & Memory & Time & 0-shot & Linear & Finetune & I2T & T2I \\
\midrule
CLIP            & 14G & 1.00$\times$ & 37.6 & 66.5 & 82.3 & 52.9 & 32.8 \\
CLIP+SimCLR   & 30G  & 2.67$\times$ &  42.8 & 72.1 & 82.6 & 58.6 & 41.3 \\
CLIP+MAE & 16G  & 1.30$\times$  &  42.1 & 68.5 & 83.2 & 57.3 & 41.1 \\
\rowcolor{Graylight} 
MaskCLIP  & 19G & 1.75$\times$  & \textbf{44.5} & \textbf{73.7} & \textbf{83.6} & \textbf{70.1} & \textbf{45.6}\\
\bottomrule
\end{tabular}}
\vspace{-3mm}
\caption{Results of boosting CLIP with different kinds of vision self-supervised learning methods. }
\vspace{-5mm}
\label{tab:target}
\end{table}

\subsection{Analysis}

We first present our analysis by studying different ways of boosting CLIP. The baseline is CLIP~\cite{radford2021learning} trained on the YFCC-15M.
Besides the introduced masked self-distillation, we consider two other popular methods: (1) SimCLR~\cite{chen2020simple}, a representative contrastive method; and (2) MAE~\cite{he2021masked} the state-of-the-art masked image modeling approaches.
All the compared methods are trained on the YFCC-15M for a fair comparison.
We have the following observations.

\noindent\textbf{Vision self-supervision helps VL contrastive.}
We evaluate the models on both vision task ImageNet-1K~\cite{deng2009imagenet} classification and vision-language task image-text retrieval on Flicker30K~\cite{young2014image} and present the comparison in Table~\ref{tab:target}.
All the added vision self-supervision, regardless of contrastive or generative, improves the baseline CLIP. Among them, our proposed MaskCLIP achieves the best results in terms of all the evaluation metrics, outperforming CLIP with +6.9\%, +7.2\%, + 1.3\% on ImageNet-1K classification for zero-shot, linear probing, and finetuning respectively, and +17.2\%, +12.8\% on Flicker30K for image-to-text retrieval and text-to-image retrieval. 
We also report the training GPU memory usage and time-consuming cost in Table~\ref{tab:target}. 
It is worth noting that the contrastive model (CLIP+SimCLR) compares two additional views of the input image, resulting in larger GPU memory usage and longer training time.

\begin{table}

\footnotesize
\centering
\resizebox{0.48\textwidth}{!}{
\setlength{\tabcolsep}{0.6mm}{
\begin{tabular}{l|c|c|c}
\toprule
\multirow{2}{*}{Method} & \multirow{2}{*}{Objective} &
ADE20K & Pascal\\
 & & mIoU   & mIoU  \\
\midrule
CLIP     & Global  &   7.2 &  13.5  \\
CLIP + SimCLR & Global + Global      & 6.3   & 11.9   \\
CLIP + MAE   & Global + Pixel-wise Local    & 8.3   & 16.4  \\
\rowcolor{Graylight} 
MaskCLIP & Global + Token-wise Local   & 10.2  & 17.2  \\

\bottomrule
\end{tabular}}}
\vspace{-3mm}
\caption{Annotation-free zero-shot segmentation results on ADE20K and Pascal Context.}
\vspace{-5mm}
\label{tab:zs-seg}
\end{table}

\noindent\textbf{Masked image modeling is able to learn representations for local patches.}
We argue that the image encoder only pays attention to the text-described objects under VL contrastive due to sparse text description and to the centric objects under image contrastive due to central-crop augmentation.
In contrast, masked image modeling forces the image encoder to focus on local patches using token-wise objectives by mandatorily masking a large portion of patches. 
Here, we provide numerical comparisons for evidence.
We conduct an ``Annotation-free zero-shot segmentation'' experiment to test the zero-shot segmentation. The results on such a dense prediction task would better reveal the ability of local patch representations than global classification. 
Following the design in DenseCLIP~\cite{zhou2021denseclip}, we use the prompted label feature as the linear classification weight to realize segmentation, without any training procedure.
We evaluate the performance on two widely used datasets: ADE20K~\cite{zhou2017scene} and Pascal Context~\cite{mottaghi2014role}. The results are shown in Table.\ref{tab:zs-seg}.
We can see that equipped with masked image modeling, our MaskCLIP as well as CLIP+MAE achieves better results than CLIP and CLIP+SimCLR, validating our hypothesis.

\noindent\textbf{Masked self-distillation learns semantic representations for local patches.}
Our masked self-distillation predicts visual features dynamically outputted by the visual encoder and thus implicitly gets supervision from the text side via VL contrastive. While MAE predicts fixed low-level pixels, making it inefficient to learn semantic representations (as the objective may force the representation to memorize low-level details) and thus causing conflict with VL contrastive. To show this,
we select images from MS-COCO~\cite{lin2014microsoftcoco} and calculate the feature similarity between image features and their corresponding caption features. 
We also select objects in the caption, prompt it to a new caption, such as ``a photo of teddy bears'', and calculate the similarities.
An example is shown in Figure~\ref{fig:vis} (More can be found in the supplementary material).
Comparing MaskCLIP with CLIP+MAE in the fourth column, we can see that CLIP+MAE uses color as evidence and fails to distinguish the white teddy bear from the white snow. While our MaskCLIP successfully differentiates the two objects, suggesting ours learn more semantic features. On the other hand, the superior results of MaskCLIP shown in Table~\ref{tab:target} and Table~\ref{tab:zs-seg} also validate this.
It is worth mentioning that CLIP and CLIP+SimCLR fail to have a correct response partition for different single objects like MaskCLIP, further justifying our second observation.

\vspace{-1mm}
\subsection{Comparison with Previous Methods}
\vspace{-1mm}
To show the effectiveness of MaskCLIP as a general vision-language pretrain method, we conduct experiments on both vision tasks and vision-language tasks. For vision tasks, we report results on ImageNet-1K~\cite{deng2009imagenet} classification, MS-COCO~\cite{lin2014microsoftcoco} object detection, and ADE20K~\cite{zhou2017scene} semantic segmentation. For vision-language tasks, we report zero-shot results on recent challenging ICinW 20 datasets benchmark and image-text retrieval results on Flickr30K~\cite{young2014image} and MS-COCO~\cite{lin2014microsoftcoco}. In the following, we compare with the supervised baseline DeiT~\cite{touvron2021training}, self-supervised methods SimCLR~\cite{chen2020simple} and MAE~\cite{he2021masked}, and vision-language methods CLIP~\cite{radford2021learning} and SLIP~\cite{mu2021slip}. For a fair comparison, we train SimCLR and MAE on YFCC-15M~\cite{thomee2016yfcc100m} with the same epochs.

\noindent\textbf{Classification on ImageNet-1K.}
As shown in Table~\ref{tab:in1k}, MaskCLIP benefits from the advantages of both VL pretraining and image mask self-distillation that shows strong performance on all the metrics. For zero-shot tasks, MaskCLIP outperforms CLIP by $+6.9\%$ with 25 epoch training and achieves $+1.7\%$ higher than the recent work SLIP. When it comes to finetune, MaskCLIP reaches $83.6\%$ top-1 accuracy, and outperforms CLIP by $+1.3\%$.

\begin{table}[t]
\centering
\footnotesize
\resizebox{0.48\textwidth}{!}{
\setlength{\tabcolsep}{0.5mm}{
\begin{tabular}{l|c|ccc|c|cc}
\toprule
\multirow{2}{*}{Method} & \multirow{2}{*}{Epoch} & 
\multicolumn{3}{c|}{IN-1K} & \multicolumn{1}{c|}{ADE20K} & \multicolumn{2}{c}{MS-COCO}\\

& & 0-Shot & Lin & FT & mIoU & AP$^b$ & AP$^m$\\
\midrule
DeiT~\cite{touvron2021training}     & 300* & -- & -- & 81.8 & 47.4 & 44.1 & 39.8\\
SimCLR~\cite{chen2020simple}        & 25   & -- & 64.0 & 82.5  & 48.0 &  44.6  & 40.2 \\
MAE~\cite{he2021masked}            & 25   & -- & 56.2 & 82.5 & 46.5 & 43.2 & 39.1 \\
CLIP~\cite{radford2021learning}         & 25   & 37.6 & 66.5  & 82.3 & 47.8 & 43.6 & 39.5\\
SLIP~\cite{mu2021slip}         & 25  & 42.8 & 72.1  & 82.6 & 48.5 &  44.0 & 40.3 \\
\rowcolor{Graylight} 
MaskCLIP          & 25   & \textbf{44.5} & \textbf{73.7} &  \textbf{83.6} & \textbf{50.5} & \textbf{45.4} & \textbf{40.9} \\


\bottomrule
\end{tabular}}}
 \vspace{-3mm}
\caption{Comparison with previous methods, including supervised baselines, self-supervised pretraining methods, and vision-language pretraining methods. * is the epoch of the supervised baseline on ImageNet-1K.
}
\vspace{-4mm}
\label{tab:in1k}
\end{table}

\noindent\textbf{Semantic segmentation on ADE20K.}
Then we apply our MaskCLIP to the semantic segmentation task. Here we use the UperNet~\cite{xiao2018unified} framework with $512\times 512$ input and end-to-end training for 160K iterations. 
The evaluation metric is the mean Intersection of Union (mIoU) and we report single-scale evaluation results here.
The results are given in Table~\ref{tab:in1k}. Our method achieves 50.5 mIoU, $+2.7$ mIoU than our baseline method CLIP, and $+2.0$ mIoU than SLIP.
This verifies the effectiveness of our introduced incorporation.

\begin{table*}[t]

\vspace{-2mm}
\setlength{\tabcolsep}{2pt}
\linespread{1}
\scriptsize
\resizebox{\textwidth}{!}{
\begin{tabular}{ll |c|cccccccccccccccccccc} 
\toprule
&
& \rotatebox[origin=lb]{90}{\smash{Average}}
& \rotatebox[origin=lb]{90}{\smash{Caltech-101}}
& \rotatebox[origin=lb]{90}{\smash{CIFAR-10}}
& \rotatebox[origin=lb]{90}{\smash{CIFAR-100}}
& \rotatebox[origin=lb]{90}{\smash{Country211}}
& \rotatebox[origin=lb]{90}{\smash{DTD}}
& \rotatebox[origin=lb]{90}{\smash{EuroSAT}}
& \rotatebox[origin=lb]{90}{\smash{FER-2013}}
& \rotatebox[origin=lb]{90}{\smash{Aircraft}} 
&\rotatebox[origin=lb]{90}{\smash{Food-101}}
& \rotatebox[origin=lb]{90}{\smash{GTSRB}}
& \rotatebox[origin=lb]{90}{\smash{Memes}}
& \rotatebox[origin=lb]{90}{\smash{KittiDis}}
& \rotatebox[origin=lb]{90}{\smash{MNIST}}
&\rotatebox[origin=lb]{90}{\smash{Flowers}}
& \rotatebox[origin=lb]{90}{\smash{Pets}}
& \rotatebox[origin=lb]{90}{\smash{PatchCam}} 
& \rotatebox[origin=lb]{90}{\smash{SST2}}
&\rotatebox[origin=lb]{90}{\smash{RESISC45}}
&\rotatebox[origin=lb]{90}{\smash{Cars}} 
& \rotatebox[origin=lb]{90}{\smash{Voc2007}} \\

\toprule
\multicolumn{20}{l}{\textit{Pretraining on YFCC-15M}} \\
\multicolumn{2}{l|}{CLIP} & 34.0 & 58.6 & 68.5 &  36.9 &  10.8 & 21.4 & 30.5 & 16.9 & 5.1 &  51.6 & 6.5 & 51.1 & 25.9 & 5.0 & 52.7 & 28.6 & 51.7 & \textbf{52.5} & 22.4 & 4.5 & 79.1
\\
\multicolumn{2}{l|}{SLIP} & 37.8 & 70.9 & \textbf{82.6} & 48.6 & 11.8 & 26.6 & 19.8 & 18.1 & 5.6 &  59.9 & \textbf{12.6} & 51.8 & 29.4 & \textbf{9.8} & 56.3 & 31.4 & \textbf{55.3} & 51.5 & 28.5 & 5.4 & 80.5
 \\
\rowcolor{Graylight} 
\multicolumn{2}{l|}{MaskCLIP} & \textbf{40.1} & \textbf{72.0} & 80.2 & \textbf{57.5} & \textbf{12.6} & \textbf{27.9} & \textbf{44.0 }
&\textbf{20.3} &\textbf{ 6.1} & \textbf{64.9} & 8.5 & \textbf{52.0} & \textbf{34.3} & 4.9 & \textbf{57.0} & \textbf{34.3} & 50.1 & 49.9 & \textbf{35.7} & \textbf{6.7} &	\textbf{82.1}
 \\
\midrule
\multicolumn{20}{l}{\textit{Pretraining on ICinW Academic Track Stting: YFCC-15M , GCC3M+12M, ImageNet-21K(ImageNet-1K is removed)}} \\
\rowcolor{Graylight} 
1st & MaskCLIP & \textbf{48.9} & 86.4 & \textbf{95.3} & \textbf{78.3} & 11.6 & \textbf{33.0} & \textbf{57.7} & 18.8 & \textbf{8.0} & \textbf{78.9} & 17.3 & \textbf{52.8} & 16.0 & 7.3 & 74.2 & \textbf{74.4} & \textbf{52.1} & 46.2 & \textbf{54.3} & \textbf{26.5} & \textbf{82.3} \\

2nd & KLITE* & 45.5 & \textbf{87.4} & 92.7& 68.8 &	8.2 & 32.2 & 27.9  & 17.4 & 4.3 & 72.4 & 11.4 & 48.4 & \textbf{31.1} & 12.8 & \textbf{75.6} &	65.9 & 50.6 & \textbf{52.9} & 44.4 &	10.2 & \textbf{82.3} \\

3rd & YT-CLIP & 44.5 & 77.8 & 83.5 & 58.4 & \textbf{11.9} & 31.9 & 40.7 & 27.1 & 6.9 & 68.7 & \textbf{18.8} & 52.3 & 9.1 & \textbf{18.8} & 53.1 & 69.3 & 51.5 & 50.3 & 52.7 & 19.7 & 79.3 \\

4th & UniCL$\dagger$  & 44.0 & 84.8 & 90.2 & 67.8 & 6.7 & 25.4 & 35.3 & \textbf{30.8} & 3.5 & 68.3 & 11.1 & 51.0 & 17.9 & 11.3 & 71.7 & 44.9 & 52.1 & 49.5 & 41.4 & 24.2 & 81.3 \\

5th & Gramer* &  43.2 & 83.9 & 92.9 & 69.5 & 7.3 & 25.5 & 24.4 & 30.4 & 2.7 & 71.0 & 9.0 & 52.6 & 12.4 & 10.1 & 70.4 & 52.4 & 50.6 & 50.1 & 44.8 & 13.8 & 81.3 \\
\bottomrule
\end{tabular}}
\vspace{-3mm}
\caption{Zero-shot evaluation on ICinW classification benchmarks. Best results in \textbf{bold}.   * indicates using Swin-B as the backbone, $\dagger$ indicates using Focal-B as the backbone. }
\vspace{-1mm}
\label{tab:downstream-linear}
\end{table*}

\begin{table*}[t]
\vspace{-3mm}
\center
\footnotesize
\resizebox{\textwidth}{!}{
\setlength{\tabcolsep}{2mm}{
\begin{tabular}{l|c|ccc|ccc|ccc|ccc}
\toprule
&&\multicolumn{6}{c|}{Flickr30K}& \multicolumn{6}{c}{MS-COCO}\\
 & Training &\multicolumn{3}{c|}{Image-to-text} & \multicolumn{3}{c|}{Text-to-image} & \multicolumn{3}{c|}{Image-to-text} & \multicolumn{3}{c}{Text-to-image} \\
& Epoch & R@1 & R@5 & R@10 & R@1 & R@5 & R@10 & R@1 & R@5 & R@10 & R@1 & R@5 & R@10 \\
\midrule
CLIP~\cite{radford2021learning}   & 25 & 52.9 & 79.6 & 87.2 & 32.8 & 60.8 & 71.2 & 27.5 & 53.5 & 65.0 & 17.7 & 38.8 & 50.5
 \\
SLIP~\cite{mu2021slip}    & 25  &  58.6 & 85.1 & 91.7  &41.3 & 68.7 & 78.6 & 33.4 & 59.8 & 70.6 & 21.5 & 44.4 & 56.3 \\
\rowcolor{Graylight} 
MaskCLIP& 25   & \textbf{70.1} & \textbf{90.3} & \textbf{95.3} & \textbf{45.6} & \textbf{73.4} &\textbf{ 82.1} & \textbf{41.4} & \textbf{67.9} &\textbf{ 77.5} & \textbf{25.5} & \textbf{49.7 }&\textbf{ 61.3} \\

\bottomrule         
\end{tabular}}}
\vspace{-3mm}
\caption{Results of zero-shot image-text retrieval on Flickr30K and MS-COCO datasets. Best results in \textbf{bold}.  }
\vspace{-3mm}
\label{tab:zero-shot-retrieval-table}
\end{table*}

\noindent\textbf{Object detection and instance segmentation on MS-COCO.}
We further investigate our transfer performance on object detection and instance segmentation in Table.\ref{tab:in1k}. Here we use Mask-RCNN~\cite{he2017mask} framework with single-scale input and $1\times$ schedule (12 epochs).
Our method achieves $45.4$ box AP and $40.9$ mask AP, $+1.8/1.4$ better than CLIP, and $+1.4/0.6$ better than SLIP.

\noindent\textbf{Zero-shot on small datasets.}
We also report zero-shot performance on 20 small datasets under the ICinW setting (see the introduction below) in Table~\ref{tab:downstream-linear}. 
We find that all the methods perform poorly on some datasets such as Aircraft(1\% acc for random guessing, we omit the description in the following), Fer(24.7\%), Country211(0.5\%), GTSRB(5.9\%), Cars(0.8\%). This might be caused by the data domain gap that the YFCC-15M contains few related images and descriptions. For the rest of the datasets, all the methods get reasonable performance and our MaskCLIP gets the best performance on most datasets.

\noindent\textbf{Image Classification in the Wild (ICinW) Challenge}
The ICinW challenge~\cite{icinw} is a newly proposed visual pretraining benchmark, which contains 20 diverse downstream classification datasets, measuring the ability of pre-training models on both the prediction accuracy and their transfer efficiency in a new task. The pretraining is limited to three datasets: YFCC-15M~\cite{thomee2016yfcc100m}, GCC3M~\cite{sharma2018conceptual}+12M~\cite{changpinyo2021conceptual} and ImageNet-21K~\cite{deng2009imagenet} (ImageNet-1K data is excluded). We pretrain our MaskCLIP on it and get the \textbf{1$_{st}$} result in the zero-shot track~\cite{icinw_zs} (we submit the results anonymously). As shown in Table~\ref{tab:downstream-linear}, the $2_{nd}$ team KLITE uses a strong Swin-B~\cite{liu2021swin} as the backbone and additional knowledge from GPT-3~\cite{brown2020language} and Wiktionary~\cite{meyer2012wiktionary}, and the $4_{th}$ use the strong Focal-B~\cite{yang2021focal} as the backbone, while our MaskCLIP greatly outperforms these methods with a simple ViT-B backbone and no additional knowledge.


\noindent\textbf{Zero-shot on text-image retrieval.}
We further report the zero-shot text-image retrieval results on two benchmark datasets, Flicr30K~\cite{young2014image} and MS-COCO~\cite{lin2014microsoftcoco}. We find that the text without any prefixes or suffixes works well for all the models. Table~\ref{tab:zero-shot-retrieval-table} shows the results. We can see that MaskCLIP exhibits a strong zero-shot performance. For example, with 25 epochs training, MaskCLIP reaches 41.4\% Rank@1 image-to-text accuracy on MS-COCO, outperforming CLIP with 13.9\%, and 25.5\% Rank@1 text-to-image accuracy, +7.8\% higher than CLIP.

\begin{table*}[h]
\vspace{-.2em}
\centering
\subfloat[
\textbf{Training Objectives ablation}. Both is necessary for MaskCLIP.
\label{tab:loss_ablation}
]{
\centering
\begin{minipage}{0.29\linewidth}{\begin{center}
\tablestyle{4pt}{1.05}
\begin{tabular}{x{36}x{24}x{24}x{36}}
\toprule
Model & 0-Shot & FT & I2T/T2I \\
\midrule
\rowcolor{Graylight} 
MaskCLIP & 44.5 & 83.6 & 70.1/45.6 \\
w/o $\mathcal{L}_{\text{MLM}}$ & 42.8 & 83.6 & 65.0/41.6\\
w/o $\mathcal{L}_{\text{Dis}}$ & 42.0 & 82.4 & 65.4/40.5\\
\bottomrule
\end{tabular}
\end{center}}\end{minipage}
}
\hspace{2em}
\subfloat[
\textbf{Distillation loss format}. The online tokenizer with cross-entropy loss works slightly better than MSE loss.
\label{tab:ce_loss}
]{
\begin{minipage}{0.29\linewidth}{\begin{center}
\tablestyle{4pt}{1.05}
\begin{tabular}{x{18}x{24}x{24}x{24}}
\toprule
Loss & 0-Shot & Lin & FT \\
\midrule
MSE & 43.8 & 73.2 & 83.6 \\
\rowcolor{Graylight} 
CE  & 44.5 & 73.7 & 83.6 \\
\multicolumn{4}{c}{~}\\
\bottomrule
\end{tabular}
\end{center}}\end{minipage}
}
\hspace{2em}
\subfloat[
\textbf{Visual decoder Depth}. A shallow decoder gets better performance.
\label{tab:image_decoder}
]{
\begin{minipage}{0.29\linewidth}{\begin{center}
\tablestyle{1pt}{1.05}
\begin{tabular}{x{28}x{24}x{24}x{24}}
\toprule
Depth & 0-Shot  & Lin & FT \\
\midrule
\rowcolor{Graylight} 
1     & 44.5 & 73.7 & 83.6 \\
2     & 43.7 & 72.9 & 83.4 \\
4     & 43.5 & 72.5 & 83.3 \\
\bottomrule
\end{tabular}
\end{center}}\end{minipage}
}
\\
\centering
\vspace{.3em}
\subfloat[
\textbf{Text decoder depth}. The decoder is necessary and a shallow one works better.
\label{tab:text_decoder}
]{
\begin{minipage}{0.29\linewidth}{\begin{center}
\tablestyle{6pt}{1.05}
\begin{tabular}{x{28}x{24}x{36}}
\toprule
Depth & 0-Shot & I2T/T2I \\
\midrule
0  & 43.5 & 65.2/44.1\\
1  & 44.3 & 70.4/45.3\\
2  & 44.3 & 70.2/45.4\\
\rowcolor{Graylight} 
4  & 44.5 & 70.1/45.6\\
8  & 44.2 & 67.5/44.7\\
\bottomrule
\end{tabular}
\end{center}}\end{minipage}
}
\hspace{2em}
\subfloat[
\textbf{Distillation loss weight}. A small loss weight works well for MaskCLIP.
\label{tab:dis_loss_weight}
]{
\begin{minipage}{0.29\linewidth}{\begin{center}
\tablestyle{1pt}{1.05}
\begin{tabular}{x{32}x{28}x{28}x{28}}
\toprule
Weight & 0-Shot  & Lin & FT \\
\midrule
1     & 38.5 & 68.2 & 82.5\\
0.1   & 44.4 & 73.5 & 83.5 \\
\rowcolor{Graylight} 
0.05  & 44.5 & 73.7 & 83.6 \\
0.01  & 43.6 & 73.0 & 83.4 \\ 
\multicolumn{3}{c}{~}\\
\bottomrule
\end{tabular}
\end{center}}\end{minipage}
}
\hspace{2em}
\subfloat[
\textbf{MLM loss weight}. A small loss weight works better.
\label{tab:mlm_loss_weight}
]{
\begin{minipage}{0.29\linewidth}{\begin{center}
\tablestyle{1pt}{1.05}
\begin{tabular}{x{36}x{28}x{40}}
\toprule
Weight & 0-Shot & I2T/T2I \\
\midrule
1    & 36.5 & 51.7/32.1\\
0.1  & 44.3 & 69.2/45.9\\
\rowcolor{Graylight} 
0.05 & 44.5 & 70.1/45.6\\
0.01 & 43.2 & 70.6/45.6\\
\multicolumn{3}{c}{~}\\
\bottomrule
\end{tabular}
\end{center}}\end{minipage}
}
\vspace{-3mm}
\caption{\textbf{MaskCLIP ablation experiments} with YFCC-15M dataset.
We report zero-shot(0-Shot), fine-tuning (FT), and linear probing (Lin) accuracy (\%) for image-encoder-related ablation. And zero shot image-to-text, text-to-image retrieval~(I2T/T2I) for text encoder-related ablations. Default settings are marked in \colorbox{baselinecolor}{gray}.}
\label{tab:ablations} 
\vspace{-6mm}
\end{table*}

\vspace{-1mm}
\subsection{Ablations}
\vspace{-1mm}
We compare our default settings with other alternatives to justify the efficacy of our model designs.

\noindent\textbf{Training objectives ablation.}
As shown in Table.\ref{tab:loss_ablation}, when we remove the mask language modeling loss $\mathcal{L}_{\text{MLM}}$, the performance of the image-text task drops, including the zero-shot accuracy and retrial performance. While benefiting from the distillation loss, the finetuning performance on ImageNet-1K is not influenced. When we remove the distillation loss $\mathcal{L}_{\text{Dis}}$, we observe a performance drop on all tasks, especially the finetuning results.

\noindent\textbf{Distillation loss format.}
Different from previous methods~\cite{dong2022boot, baevski2022data2vec, he2021masked} that calculate the per-element distance as the loss function, we use an online tokenizer to map the feature to soft codewords and use the cross-entropy loss as the supervision. Here we study their difference in Table.\ref{tab:ce_loss}. We find that although they get similar fine-tuning performance, the CE loss gets better zero-shot and linear probing performance. 
The reason may be that the per-element MSE loss leads the model to fit some unnecessary details of the target feature, while the CE loss with soft tokenizer  helps the model to focus more on the important feature.

\noindent\textbf{Distillation \& MLM loss weight.}
Here we set the loss weight of the CLIP branch as 1 and study the loss weight of the two additional branches. As shown in Table.\ref{tab:dis_loss_weight} and Table.\ref{tab:mlm_loss_weight}, setting $\lambda=1$ or $\beta=1$ emphasize too much on new tasks, which mislead 
the model to a wrong converge direction, resulting in poor performance. When we reduce the loss weight by $10\times$, the two additional tasks are helpful for the model and show a consistent gain on all the metrics. We suspect this is because the CLIP loss requires two different capabilities: understanding the input content and aligning them into a shared feature space. And the goal of the two additional self-supervised learning tasks is to facilitate understanding.

\noindent\textbf{Image \& Text decoder depth.}
Then we study the influence of the decoder depth for both image and text decoders. As shown in Table.\ref{tab:image_decoder}, we find the image decoder with only one layer works well, increasing the decoder depth leads to worse performance on all metrics.
Similarly, Table.\ref{tab:text_decoder} shows that the text branch benefits from a shallow decoder design. We argue that a too-deep decoder would make the encoder lazy, relying on the strong decoder to resolve the challenging mask feature/language modeling tasks. And the different depth choice between the image and text branches is caused by the framework difference: the image branch sees the mask tokens at the decoder, while the text branch takes the mask tokens as the encoder input.
Note that if we remove the text decoder, the performance gets worse. We think this is largely caused by the output conflict that the global recognition feature aggregation and local word prediction are conducted at the same layer.

\noindent\textbf{Single-Stage v.s. two-Stage.}
Our MaskCLIP learns the VL contrastive and masked self-distillation simultaneously and jointly in a single stage. One possible variant is to first train CLIP and then use CLIP feature from the first stage to train masked image modeling as in~\cite{wei2021masked,wei2022mvp}.
We report results on three datasets in Table~\ref{tab:stage}.
We can see that the second stage achieves better finetuning results compared with results from the stage one, showing the effectiveness of masked image modeling.
Nonetheless, such two-stage training requires longer training time and loses the transfer capability in a zero-shot setting. In contrast, our MaskCLIP achieves superior results under all settings with fewer epochs.

\begin{table}
\centering
\footnotesize
\resizebox{0.48\textwidth}{!}{
\setlength{\tabcolsep}{0.4mm}{
\begin{tabular}{l|c|c|cc|cc|cc}
\toprule
\multicolumn{2}{c|}{\multirow{2}{*}{Method}} &
\multirow{2}{*}{Epoch} & \multicolumn{2}{c|}{IN-1K} & \multicolumn{2}{c|}{Flicker30K} & \multicolumn{2}{c}{ADE20K}\\
\multicolumn{2}{c|}{ } &  & 0-shot  & FT & I2T & T2I & 0-shot  & FT  \\
\midrule
\multirow{3}{*}{ Two-Stage} & Stage1 & 25   & 37.6  & 82.3  & 52.9 & 32.8 & 7.2 & 47.8 \\
\cmidrule(rl){2-9}
  & Stage2  & 25  & ---   & 83.4& --- & ---& --- & 48.2 \\
\midrule
\rowcolor{Graylight} 
\multicolumn{2}{c|}{MaskCLIP} & 25  & \textbf{44.5} &  \textbf{83.6} & \textbf{70.1} & \textbf{45.6} & \textbf{10.2} & \textbf{50.5}  \\
\bottomrule
\end{tabular}}}
\vspace{-3mm}
\caption{Comparison between two-stage method and our single-stage MaskCLIP.}
\vspace{-4mm}
\label{tab:stage}
\end{table}

\vspace{-2mm}
\section{Conclusion}
\vspace{-1mm}
We present MaskCLIP, a new VL pretraining framework that incorporates masked self-distillation into VL contrastive. 
We point out that masked self-distillation learns local semantics, fitting nicely to the VL contrastive that aims to learn global semantics, and this is supported with comprehensively designed experiments.
We also utilize mask language modeling to enhance the text encoder which is critical for zero-shot performance.
The resulting visual encoder shows strong transfer capability across widely adopted benchmarks for linear probing, fine-tuning, and also zero-shot evaluation.


{\small
\bibliographystyle{ieee_fullname}
\bibliography{egbib}
}

\clearpage
\appendix

\section{More Experiment}

\noindent\textbf{Comparison over small model and small dataset.}
As some baselines report ViT-B/32 instead of ViT-B/16, in order to compare,
we further experiment MaskCLIP with a smaller model ViT-B/32 and report the zero-shot performance on ImageNet-1K.
As shown in Table~\ref{tab:ablations} left, our MaskCLIP outperforms the combination~\cite{cui2022democratizing} of two recent strong methods DeCLIP~\cite{li2022supervision} and FILIP~\cite{yao2021filip}.
We also investigate the performance on a smaller dataset CC3M~\cite{sharma2018conceptual} (we use ViT-B/16 here in coherency with previous experiments). Table~\ref{tab:ablations}(right part) shows that MaskCLIP achieves consistent gain.

\begin{table}[h]
\centering
\small 
\renewcommand\arraystretch{0.6}
\setlength{\tabcolsep}{1.1mm}{
\begin{tabular}{l|ccc|l|ccc}
\toprule
ViT-B/32 & 0-Shot & Lin & FT & CC3M & 0-Shot & Lin & FT \\
\midrule
CLIP     & 26.1 & 60.5 & 74.3 & CLIP     & 17.1 & 53.3 & 78.5 \\
DeFILIP  & 36.4 & --- & --- & SLIP     & 23.0 & 65.4 & 81.4 \\
MaskCLIP & 38.5 & 69.1 & 79.2 & MaskCLIP & 24.4 & 66.1 & 82.5 \\

\bottomrule
\end{tabular}}
\caption{Results of zero-shot performance on ImageNet-1K when pretrained with  ViT-B/32 model(left) or CC3M dataset(right) .}
\label{tab:ablations} 
\end{table}
 
\noindent\textbf{Ablation on distillation loss.}
Here we further study the effectiveness of each component in the distillation loss. We start from CLIP+MAE and add three components of the distillation loss one by one. We find that 1) using the feature as the prediction target improves all metrics; 2) using EMA model gets better performance; 3) the MLM loss improves all the vision-language tasks.

\begin{table}[h]
\centering
\small 
\renewcommand\arraystretch{0.6}
\setlength{\tabcolsep}{1.1mm}{
\begin{tabular}{l|cc|c|c|c}
\toprule
 & 0-Shot & FT & Seg & Det & I2T/T2I \\
\midrule
CLIP+MAE (baseline)  & 42.1 & 83.2 & 49.1 & 44.5/40.4 & 57.3/41.1\\
+ Feature prediction & 42.6 & 83.4 & 49.9 & 45.1/40.6 & 62.3/41.4 \\
+ EMA model          & 42.8 & 83.6 & 50.4 & 45.5/40.9 & 65.0/41.6\\
+ MLM loss           & 44.5 & 83.6 & 50.5 & 45.4/40.9 & 70.1/45.6 \\

\bottomrule
\end{tabular}}
\caption{Component ablation of the distillation loss.}
\vspace{-5mm}
\end{table}

\section{Experiment detail}
\textbf{Pre-training}
We train our proposed MaskCLIP from scratch and training for 25 epochs, the batch size is fixed to 4096 for all the experiments. We use 32 V100 for training with 128 samples per GPU.
We use the AdamW~\cite{loshchilov2017decoupled} optimizer with weight decay 0.1. The learning rate is set to $1e^{-3}$ with one epoch warm-up and decay to $1e^{-5}$ followed by a cosine schedule. 
The masks used in the mask self-distillation branch are random mask with a mask ratio of 75\%. 
The EMA weight is set to 0.999 and linearly increases to 0.9999 during the training.
We pretrain all the models with the commonly used YFCC15M dataset, which is flited from the YFCC100M~\cite{thomee2016yfcc100m} dataset by ~\cite{radford2021learning}. 

For the ICinW academic track experiment, we pretrain the model with three datasets: YFCC-15M~\cite{thomee2016yfcc100m}, GCC3M~\cite{sharma2018conceptual}+12M~\cite{changpinyo2021conceptual} and ImageNet-21K~\cite{deng2009imagenet} (ImageNet-1K data is excluded). Here we use the UniCL~\cite{yang2022unified} to utilize the ImageNet-22k dataset in the pretraining with a unified format. We train the model for 32 epochs and 16384 batch size, the rest settings are the same as the YFCC15M setting. 

\textbf{Zero-shot ImageNet-1K classification. }
For zero-shot on ImageNet-1K, we follow the prompt setting in ~\cite{mu2021slip} to convert the labels to text features, which contains 7 prompt templates and we use the average feature as the final label feature. We calculate the similarity between image feature and all the label features to get its zero-shot classification result.

\textbf{Linear-probing ImageNet-1K classification. }
For linear probing, we fix the backbone and train a new linear classifier for 90 epochs. Following the setting in MAE~\cite{he2021masked}, we add a batch-norm layer without learnable affine parameters before the classifier to avoid adjusting the learning rate for each model. We set the batch size to 16384 and use the LARS~\cite{you2017lars} optimizer with weight decay 0 and momentum 0.9. The learning rate is set to 6.4 and decays to 0 following the cosine schedule.

\textbf{Fine-tuning ImageNet-1K classification. }
When fine-tuning on the ImageNet-1K dataset, we average pool the output of the last transformer of the encoder and feed it to a softmax-normalized classifier.
We fine-tune 100 epochs for all the experiments, the learning rate is warmed up to 0.0006 for 20 epochs and decay to $1e^{-6}$ following the cosine schedule. 
Similar to recent works, we also apply the layer decayed learning rate used in ~\cite{bao2021beit} and we set the decay factor as 0.7. Note that we use the pure ViT architecture, \emph{without} the techniques used in  ~\cite{bao2021beit}, such as layer scale and relative position embedding.
The evaluation metric is top-1 validation accuracy of a single $224\times 224$ crop.

\noindent \textbf{Zero-shot Semantic segmentation.}
Here we follow the setting in DenseCLIP~\cite{zhou2021denseclip} based on the implementation from mmsegmentaion~\cite{mmseg2020}. 
For ADE20K and MS-COCO, we report the single-scale test result with $512\times512$ input. For Pascal Context, we use $480\times480$ input. To avoid the influence of position embedding caused by changing input size, we use sliding inference with $224\times224$ input and stride $112$.
To convert the labels to text embedding, we use 85 prompt templates and use the average feature as the final label feature.

\noindent \textbf{ADE20K Semantic segmentation.}
Here we use: UperNet~\cite{xiao2018unified} based on the implementation from mmsegmentaion~\cite{mmseg2020}. 
For UperNet, we follow the settings in ~\cite{bao2021beit} and use AdamW~\cite{loshchilov2017decoupled} optimizer with initial learning rate $2e^{-4}$, weight decay of 0.05 and batch size of 16 (8 GPUs with 2 images per GPU) for 160K iterations. The learning rate warmups with 1500 iterations at the beginning and decays with a linear decay strategy. We use the layer decay ~\cite{bao2021beit} for the backbone and we set it as 0.6. 
As the ViT architecture outputs features with the same size, here we add four different scale FPNs to scale the feature map into different size. Specifically, we upsample the output feature of the $4th$ block $4\times$, upsample the output feature of the $6th$ block $2\times$, keep the output feature of the $8th$ block unchanged and downsample the output feature of the $12th$ block $2\times$. 
We use the default augmentation setting in mmsegmentation including random horizontal flipping, random re-scaling (ratio range [0.5, 2.0]) and random photo-metric distortion. All the models are trained with input size $512\times512$. The stochastic depth is set to 0.1. When it comes to testing, we report single-scale test result.

\noindent \textbf{COCO Object Detection and Instance Segmentation.}
We use the classical object detection framework Mask R-CNN~\cite{he2017mask} based on the implementation from mmdetection~\cite{mmdetection}. We train it the $1\times$ schedule with single-scale input (image is resized so that the shorter side is 800 pixels, while the longer side does not exceed 1333 pixels) for 12 epochs. We use AdamW~\cite{loshchilov2017decoupled} optimizer with a learning rate of $1e^{-4}$, weight decay of 0.05 and batch size of 16. We also use the layer decay ~\cite{bao2021beit} for the backbone and we set it as 0.75. 
The learning rate declines at the $8th$ and $11th$ epoch with decay rate being 0.1. The stochastic depth is set to 0.1. 
Similar to the implementation of semantic segmentation above, we also use four different scale FPNs to scale the feature map into different size.

\section{More visualization results.}
Here we provide more visualization results on the MS-COCO val set. In most cases, our MaskCLIP gets a better feature alignment performance between image and text.

\section{Societal impacts}
MaskCLIP is an improvement of CLIP, so it has the same societal impacts of CLIP, including some malicious usages and positive applications. Meanwhile, CLIP and MaskCLIP may suffer from some unwanted data bias, as the data used for training are roughly collected from the Internet.

\begin{figure*}
\centering
\includegraphics[width=0.85\linewidth]{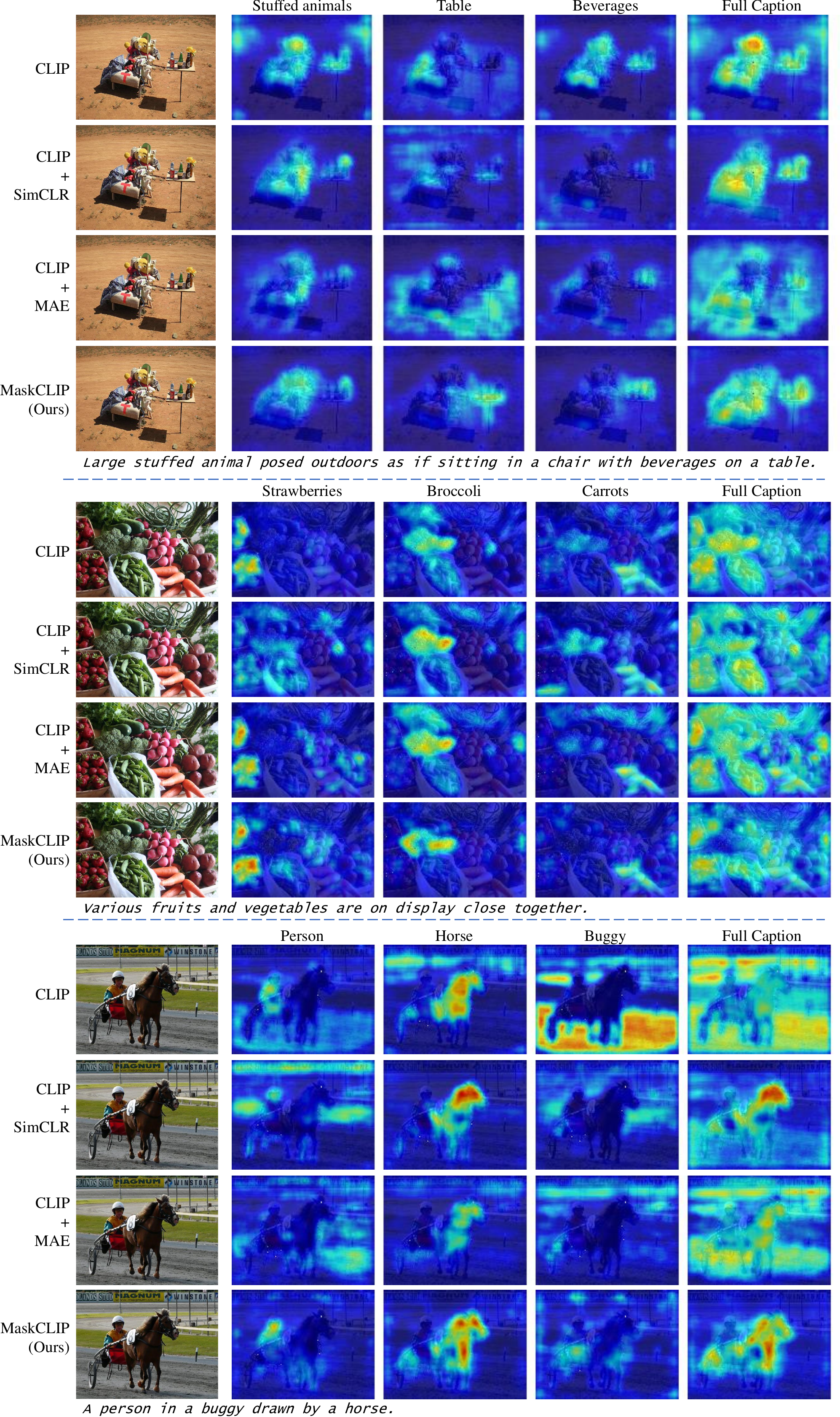}

\captionof{figure}{Visualization of the similarity between text and image features. The images and captions are from the MS-COCO val set.} 
\label{fig:vis}
\vspace{-5mm}
\end{figure*}

\begin{figure*}
\centering
\includegraphics[width=1\linewidth]{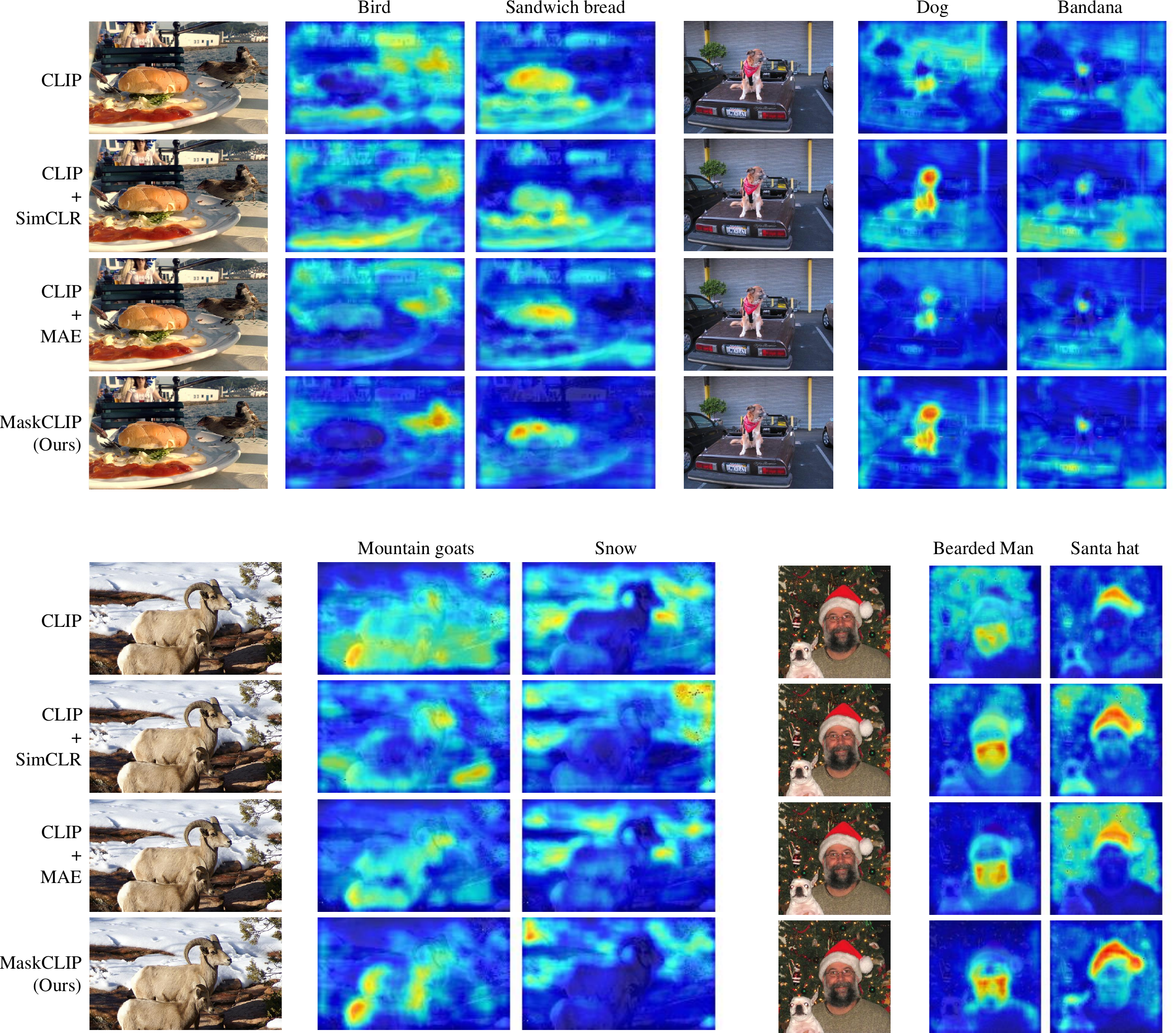}

\captionof{figure}{Visualization of the similarity between words and image features. The images and captions are from the MS-COCO val set.} 
\label{fig:vis}

\end{figure*}

\begin{figure*}
\centering
\includegraphics[width=1\linewidth]{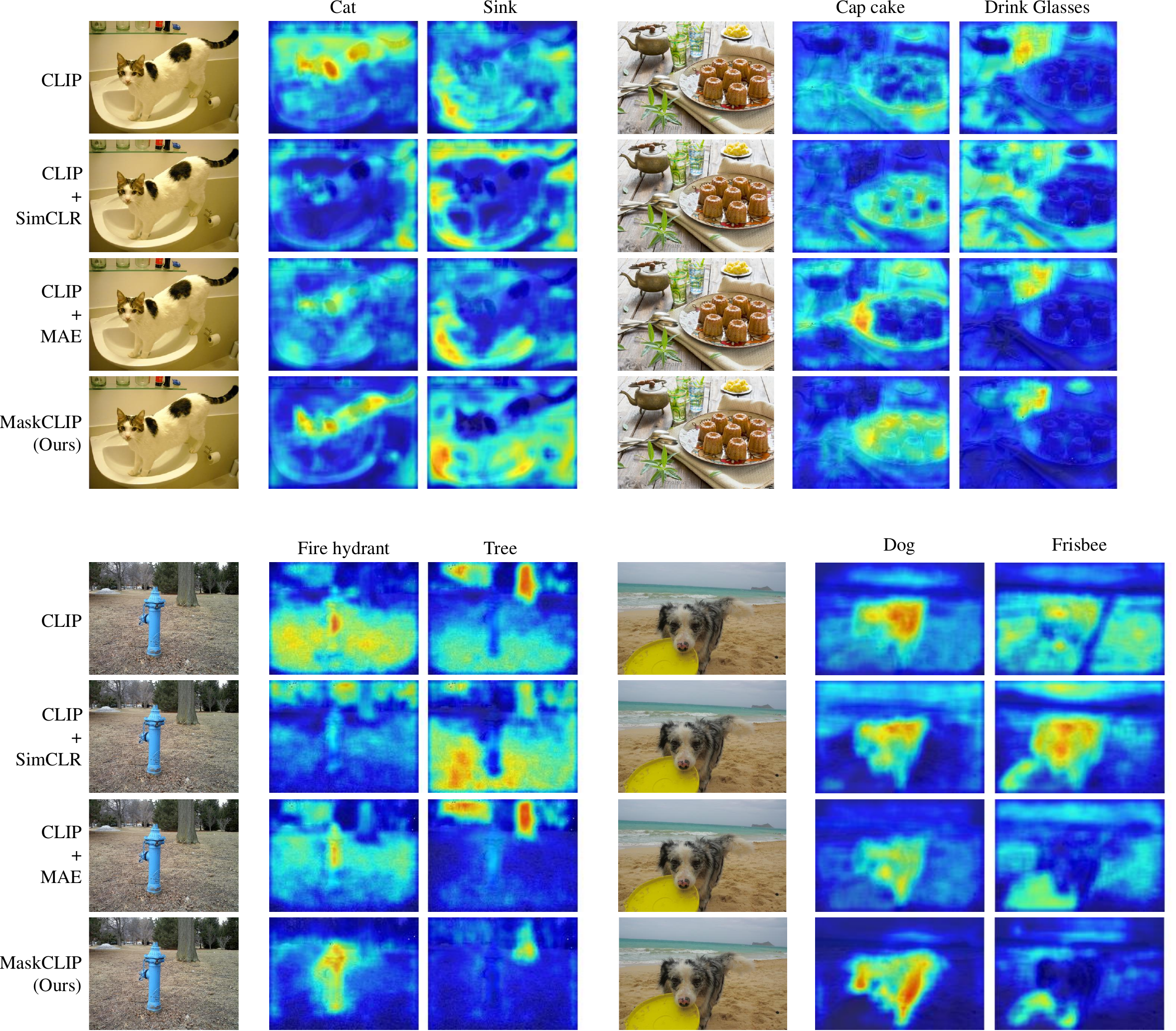}

\captionof{figure}{Visualization of the similarity between words and image features. The images and captions are from the MS-COCO val set.} 
\label{fig:vis} 
\end{figure*}

\end{document}